\documentclass{llncs}
\pdfoutput=1
\usepackage{tabularx}
\usepackage{array}
\usepackage{paralist}
\usepackage{url}
\usepackage{pbox}
\usepackage{graphicx}

\usepackage[numbers,sort,sectionbib]{natbib}

\usepackage{tikz} % for COD example

\newcommand{\codiag}{\textit{C-O Diagram}}
\newcommand{\codiags}{\textit{C-O Diagrams}}
\newcommand{\UPPAAL}{\textsc{Uppaal}}

\newcommand{\name}[1]{{\sffamily\small #1}}
\newcommand{\mname}[1]{\ensuremath{\mathsf{#1}}}
\newcommand{\isPred}[2]{\ensuremath{\mathit{is{#1}}(\mname{#2})}}
\newcommand{\isDone}[1]{\isPred{Done}{#1}}
\newcommand{\rep}[1]{\ensuremath{\#\mname{#1}}}

\newcolumntype{L}{>{\raggedright\arraybackslash}X}%

\begin{document}

\pagestyle{headings} % switches on printing of running heads

\mainmatter % start of the contributions

\title{Extracting Formal Models from Normative Texts}
\titlerunning{Extracting Formal Models from Normative Texts}

\author{John J. Camilleri\inst{1} \and Normunds Gruzitis\inst{2} \and Gerardo Schneider\inst{1}}
\authorrunning{Camilleri et al.}

\institute{
Department of Computer Science and Engineering,\\
Chalmers University of Technology and University of Gothenburg, Sweden
\and
Institute of Mathematics and Computer Science, University of Latvia
\email{\{john.j.camilleri,gerardo\}@cse.gu.se}, \email{normunds.gruzitis@lu.lv}
}

\maketitle

\begin{abstract}
We are concerned with the analysis of \emph{normative texts}---documents based on the deontic notions of obligation, permission, and prohibition.
Our goal is to make queries about these notions and verify that a text satisfies certain properties concerning causality of actions and timing constraints.
This requires taking the original text and building a representation (model) of it in a formal language, in our case the \codiag{} formalism.
%This ``translation'' has so far been a manual task, requiring the user to read a text sentence by sentence and encode it as a \codiag.
We present an experimental, semi-automatic aid that helps to bridge the gap between a normative text in natural language and its \codiag{} representation.
Our approach consists of using dependency structures obtained from the state-of-the-art Stanford Parser, and applying our own rules and heuristics in order to extract the relevant components.
The result is a tabular data structure where each sentence is split into suitable fields, which can then be converted into a \codiag.
The process is not fully automatic however, and some post-editing is generally required of the user.
We apply our tool and perform experiments on documents from different domains, and report an initial evaluation of the accuracy and feasibility of our approach.
\end{abstract}

\section{Introduction}

\begin{figure}
    \centering
    \scalebox{0.75}{\begin{tikzpicture}
  \tikzstyle{block} = [rectangle, draw, text centered, align=center]
  \tikzstyle{circ} = [circle, draw, fill=white]
  \newcommand{\codbox}[8]{
    % x, y, agent, guard, interval, mod, rep, name
    \node at (#1-0.7,#2+1) {#3}; % agent
    \node [block,minimum height=0.75cm,minimum width=2.5cm] at (#1-1.85,#2+0.375) {#4}; % guard
    \node [block,minimum height=0.75cm,minimum width=2.5cm] at (#1-1.85,#2-0.375) {#5}; % interval
    \node [block,minimum height=1.5cm,minimum width=2cm] at (#1+0.4,#2) {#6}; % modality
    \node [block,minimum height=1.5cm,minimum width=1.4cm] at (#1+2.1,#2) {#7}; % reparation
    \node at (#1+1.3,#2-1) {#8}; % name
  }

  \codbox{0}{0}
    {}
    {$\isDone{request}$}
    {}
    {}
    {$\rep{credit}$}
    {$\mname{respond}$}
  \codbox{-3.5}{-4}
    {$\mname{company}$}
    {$\isDone{SLA1}$}
    {$t_\mathit{respond1}<24$}
    {Obligation\\$\mname{respond}$}
    {$\bot$}
    {$\mname{respond1}$}
  \codbox{+4.5}{-4}
    {$\mname{company}$}
    {$\isDone{SLA2}$}
    {$t_\mathit{respond2}<4$}
    {Obligation\\$\mname{respond}$}
    {$\bot$}
    {$\mname{respond2}$}

  \draw (0.4,-0.75) -- (0.4,-2) -- (-3.1,-2) -- (-3.1, -3.25);
  \draw (0.4,-0.75) -- (0.4,-2) -- (+4.9,-2) -- (+4.9, -3.25);
  \node [circ] at (0.4,-2) {AND};

\end{tikzpicture}}
    \caption{Example of a \codiag, showing a clause which is refined as a conjunction of obligations where \name{company} must \name{respond} within an amount of time dependant on external clauses \name{SLA1} and \name{SLA2}.}
    \label{fig:codiag}
\end{figure}
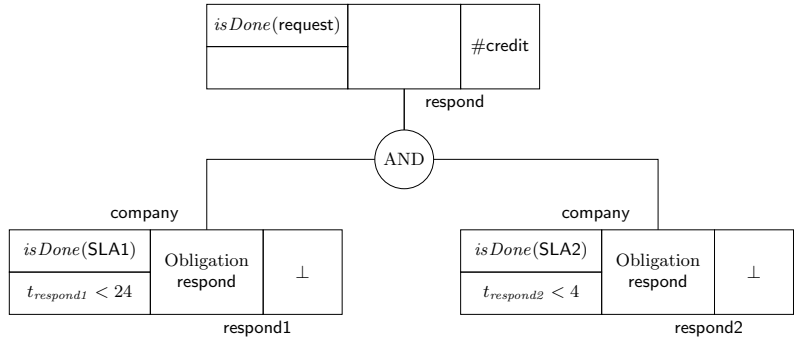

\emph{Normative texts} are natural language documents which are concerned with what must be done, may be done, or should not be done---also known as \emph{deontic norms}.
This class of documents often includes legal contracts, terms of services, regulations and service level agreements.
Our work involves the analysis of such texts using formal methods.
We achieve this by modelling normative documents within a formalism that then allows us to perform complex queries and verify properties about them.
The formalism used for this task is \codiags{} \cite{MCD+10mvs,Diaz2014},
which provides a language for visualising normative texts involving the modalities of obligation, permission and prohibition (forbiddance).
These are indicated by the letters \textbf{O}, \textbf{P} and \textbf{F} respectively.
It allows the expression of these norms over different agents and actions, together with {\it reparations} which apply when obligations and prohibitions are violated.
Figure~\ref{fig:codiag} shows an example \codiag.

Models in this formalism can be converted in an automatic, deterministic way into networks of timed automata \cite{Alur1994,Bengtsson2004},
which are amenable to verification using the \UPPAAL{} model checker \cite{Behrmann2006}.
There is, however, a large gap between the natural language texts as written by humans, and the formal representation used for automated analysis.
Because of this, the task of modelling a text is completely manual, requiring a good knowledge of both the domain and the formalism.
In this paper we present a method which helps to bridge this gap, by automatically extracting a partial model using NLP techniques.

\paragraph{Contributions}

We present here our technique for processing normative texts written in natural language and building partial models from them by analysing their syntactic structure and extracting relevant information.

Our method uses dependency structures obtained from a general-purpose statistical parser, namely the Stanford Parser \cite{Klein03,Socher13}, which are then processed using custom rules and heuristics that we have specified based on a small development corpus in order to produce a table of clauses (draft predicate candidates).
This can be seen as a specific information extraction task.
While this method may only produce a partial model, requiring further post-editing by the user, we aim to save the most tedious work so that the user (knowledge engineer) can focus better on formalisation details.

We discuss the application of this method to a small test corpus of unseen sentences, and report on the performance based on a precision-recall metric.

%% Potentially, an interactive editor for contracts facilitating the clarity and precision of text, unveiling syntactic and semantic ambiguity, redundancy and inconsistency.
%% Providing a more consistent linguistics analysis of the input text compared to a purely manual approach.

%% ---------------------------------------------------------------------------

\section{Extracting Draft Predicate Candidates}

The proposed approach is application-specific but domain-independent, assuming that normative texts (or \emph{contracts}), tend to follow a certain restricted style of natural language, even though there are variations across and within domains.
However, we do not impose any grammatical or lexical restrictions on the input texts, therefore we first apply a general-purpose parser acquiring a syntactic dependency tree representation for each  sentence.
Provided that the syntactic analysis does not contain significant errors, we then apply a number of interpretation rules and heuristics on top of dependency structures.
If the extraction is successful, one or more predicate candidates are acquired for each input sentence as shown in Table~\ref{tab:sample-output}.
More than one candidate is extracted in case of explicit or implicit coordination of subjects, verbs, objects or main clauses.

\begin{table}[]
\small
\caption{Sample input and partial output.}
\label{tab:sample-output}
\medskip
\begin{tabularx}{\textwidth}{
	>{\hsize=0.48\hsize}L|	% 8% of 6\hsize	Refin.
	>{\hsize=0.42\hsize}L|	% 7% of 6\hsize	Mod.
   	>{\hsize=0.84\hsize}L|	% 14% of 6\hsize	Subject
   	>{\hsize=1.26\hsize}L|	% 21% of 6\hsize	Verb
	>{\hsize=1.56\hsize}L|	% 26% of 6\hsize	Object
	>{\hsize=1.44\hsize}L	% 24% of 6\hsize	Modifiers
	% sum = 6.0\hsize for 6 columns
}
\cline{1-6}
\multicolumn{1}{l}{\textbf{Refin}.} & \multicolumn{1}{l}{\textbf{Mod}.} & \multicolumn{1}{l}{\textbf{Subject} (S)} & \multicolumn{1}{l}{\textbf{Verb} (V)} & \multicolumn{1}{l}{\textbf{Object} (O)} & \multicolumn{1}{l}{\textbf{Modifiers}} \\
\cline{1-6}
\multicolumn{6}{p{\textwidth}}{\textbf{1.} \textit{You must not, in the use of the Service, violate any laws in your jurisdiction (including but not limited to copyright or trademark laws).}} \\
\cline{1-6}
& F & User & violate & law & V: in User's jurisdiction \newline V: in the use of the Service \\
\cline{1-6}
\multicolumn{6}{p{\textwidth}}{\textbf{2.} \textit{You will not post unauthorized commercial communication (such as spam) on Facebook.}} \\
\cline{1-6}
& F & User & post & unauthorized commercial communication & O: such as spam \newline O: on Facebook \\
% 'on Facebook': the parser made a PP attachment error (communications on Facebook)
\cline{1-6}
\multicolumn{6}{p{\textwidth}}{\textbf{3.} \textit{You will not upload viruses or other malicious code.}} \\
\cline{1-6}
& F & User & upload & virus &  \\
\cline{2-6}
OR & F & User & upload & other malicious code &  \\
\cline{1-6}
\multicolumn{6}{p{\textwidth}}{\textbf{4.} \textit{Your login may only be used by one person - a single login shared by multiple people is not permitted.}} \\
\cline{1-6}
& P & person & use & login of User & S: one \\
\cline{1-6}
\multicolumn{6}{p{\textwidth}}{\textbf{5.} \textit{The renter shall pay all reasonable attorney and other fees, the expenses and costs incurred by owner in protection its rights under this rental agreement and for any action taken owner to collect any amounts due the owner under this rental agreement.}} \\
\cline{1-6}
& O & renter & pay & reasonable attorney & V: under this rental agreement \\
\cline{2-6}
AND & O & renter & pay & other fee & V: under this rental agreement \\
\cline{1-6}
\multicolumn{6}{p{\textwidth}}{\textbf{6.} \textit{The equipment shall be delivered to renter and returned to owner at the renter's risk, cost and expense.}} \\
\cline{1-6}
& O & equipment & [is] delivered [to] & renter & V: at renter's risk, cost and expense \\
\cline{2-6}
AND & O & equipment & [is] returned [to] & owner & V: at renter's risk, cost and expense \\
\cline{1-6}
\end{tabularx}
\end{table}

The dependency representation instead of phrase-structure representation allows for a more straightforward predicate extraction based on the syntactic relations instead of phrase types of the parts of a sentence.

In our experiment, we use the Stanford Parser whose accuracy on Penn Treebank (the WSJ section) is around 90\% \cite{Socher13}.
The Stanford dependency representation \cite{deMarneffe:2008} is being increasingly adapted to parsers for other languages as well, for instance, Chinese \cite{Chang2009}, Finnish \cite{TurkuTreebank} and Persian \cite{SERAJI14.378}, and it is the basis for the Universal Dependencies project \cite{deMarneffe:2014}.
However, our approach as such is not restricted to the specific parser or dependency representation.

\subsection{Expected Input and Intended Output}

The basic requirement for pre-processing the input text is that it is split by sentence and that only the relevant sentences are included.
In this experiment, we have manually selected the relevant sentences, ignoring (sub)titles, introductory notes etc.
Automatic analysis of the document structure is a separate issue.
We also expect that sentences do not contain grammatical errors that would considerably affect the syntactic analysis and thus the output of our tool.
The output is a table (in tab-separated format) where each row corresponds to a \codiag\ box (clause), containing fields for:
\begin{description}
  \item[Subject] the agent of the clause;
  \item[Verb] the verbal component of an action;
  \item[Object] the object component of an action (optional);
  \item[Modality] obligation (O), permission (P), prohibition (F), or declaration (D) for clauses which only state facts;
  \item[Refinement] whether a clause should be attached to the preceding clause by conjunction (AND), choice (OR) or sequence (SEQ); % NG: do verb modifiers like ``then'' indicate SEQ (e.g. in ``It shall then be possible...'')? I haven't consider this operator so far...
  \item[Time] adverbial modifiers clearly indicating temporality;
  \item[Adverbials] other adverbial phrases that modify the action; % NG: they all should be attached to Verb
  \item[Conditions] phrases indicating conditions on agents, actions or objects; % and the element they attach to
  \item[Notes] other phrases that provide additional information (e.g. relative clauses), indicating the element (head word) they attach to.
\end{description}

Values of the Subject, Verb and Object fields undergo certain normalisation and formatting: head words are lemmatised; Saxon genitives are converted to of-constructions if contextually possible; the preposition ``to'' is explicitly added to indirect objects; prepositions of prepositional objects are included in the Verb field as part of the predicate name, as well as the copula if the predicate is expressed by a participle, adjective or noun; definite and indefinite articles are omitted.

A complete document in this format can be converted automatically into a \codiag{} model. % TODO (?): a section at the end illustrating a use case
Our tool however does not necessarily produce a \emph{complete} table, in that fields may be left blank when we cannot determine what to use.
There is also the question of what is considered \emph{correct} output.
It may be the case that certain clauses may be encoded in multiple ways, and, while all fields may be filled, the user may find it more desirable to change the encoding.

\subsection{Rules}
\label{sec:rules}

\begin{figure}
    \centering
    \includegraphics[width=0.8\textwidth]{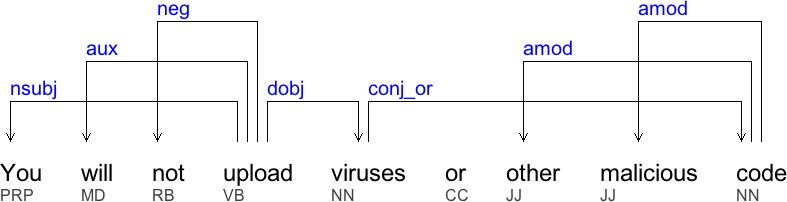}
    \caption{Sample dependency tree.}
    \label{fig:parse}
\end{figure}

We make a distinction between rules and heuristics that are applied on top of Stanford Dependencies.
Rules are everything that explicitly follow from the dependency relations and part-of-speech tags. For example, the head of the subject noun phrase (NP) is labelled by \texttt{nsubj}, and the head of the direct object NP---by \texttt{dobj} (see Figure~\ref{fig:parse}); fields Subject and Object of the output table can be straightforwardly populated by the respective phrases (as in Table~\ref{tab:sample-output}).

We also count as lexicalised rules cases when the decision can be obviously made by considering both the dependency label and the head word.
For example, modal verbs and other auxiliaries of the main verb are labelled as \texttt{aux} but words like ``may'' and ``must'' clearly indicate the respective modality (P and O).
Auxiliaries can be combined with other modifiers, for example, the modifier ``not'' (\texttt{neg}) which indicates prohibition.
In such cases, the rule is that obligation overrides permission, and prohibition overrides both obligation and permission.

In order to provide concise values (terms) for the Subject and Object fields, relative clauses (\texttt{rcmod}), verbal modifiers (\texttt{vmod}) and prepositional modifiers (\texttt{prep}) that modify heads of the subject and object NPs are separated in the Notes field.
Adverbial modifiers (\texttt{advmod}), prepositional modifiers and adverbial clauses (\texttt{advcl}) that modify the main verb are separated, by default, in the Adverbials field.

If the main clause is expressed in the passive voice, and the agent is mentioned (expressed by the preposition ``by''), the resulting predicate is converted to the active voice (as shown by the fourth example in Table~\ref{tab:sample-output}). % NG: First thing TODO after submitting the paper

% TODO: Describe copying of Subject/Object/Adverbials among the refined predicates

\subsection{Heuristics}
\label{sec:heuristics}

In addition to the obvious extraction rules, we apply a number of heuristic rules based on the development examples and our intuition about the application domains and the language of normative texts.

First of all, auxiliaries are compared and classified against extended lists of keywords.
For example, the modal verb ``can'' most likely indicates permission while ``shall'' and ``will'' indicate obligation.
In addition to auxiliaries, we also consider the predicate itself (expressed by a verb, adjective or noun). For example, words like ``responsible'', ``liable'' and ``require'' most likely express obligation.

For prepositional phrases (PP) that are direct dependants of Verb, we first check if they reliably indicate a temporal modifier and thus should be put in the Time field.
The list of such prepositions include ``after'', ``before'', ``until'', ``during'' etc.
If the preposition is ambiguous, the head of the noun phrase is checked if it bears a meaning of time.
There is a relatively open list of such potential keywords, including ``day'', ``week'', ``month'' etc.
Due to PP-attachment errors that syntactic parses often make, if a PP is attached to Object, and it has the above mentioned indicators of a temporal meaning, the phrase is put in the Verb-dependent Time field.

Similarly, we check the markers (\texttt{mark}) of adverbial clauses if they indicate time (``while'', ``when'' etc.) or a condition (e.g. ``if''), as well as values of simple adverbial modifiers, looking for ``always'', ``immediately'', ``before'' etc.
Adverbial modifiers are also checked against a list of irrelevant adverbs used for emphasis (e.g. ``very'') or as gluing words (e.g. ``however'', ``also").

NPs of the Subject and Object fields are checked for attributes: if an NP is modified by a number (\texttt{num} or \texttt{number}), these modifiers are treated as conditions and are separated in the respective field.

If there is no direct object in the sentence, or, in the case of the passive voice, no agent expressed by a prepositional phrase (using the preposition ``by''), the first PP governed by Verb is treated as a prepositional object and thus is included in the Object field.
Indirect objects result in additional refinement predicates.
% First thing TODO: adjust the implementation accordingly

Additionally, anaphoric references by personal pronouns are detected, normalised and tagged (e.g. ``we'', ``our'' and ``us'' are all rewritten as ``$<$we$>$'').
In the case of terms of services, for instance, pronouns ``we'' and ``you'' are often used to refer to the service and the user respectively.
The tool can be customised to do such a simple but effective anaphora resolution (see Table~\ref{tab:sample-output}).

\subsection{Post-editing}

Since we do not intend for our tool to be a complete replacement for a human knowledge engineer, a certain amount of post-editing is often required.
This post-editing can be categorised into the following different types, listed here in approximate order of effort required:

\begin{compactenum}
\item Filling in empty fields.
\item Adding or removing adverbial information from the subject and object.
\item Changing the verb or modality.
\item Refinement into sub-clauses.
\item Complete paraphrasing.
\end{compactenum}

%% ---------------------------------------------------------------------------

\section{Experiments}

In order to test the potential and feasibility of the proposed approach, we have selected four normative texts from three different domains:
two terms of service agreements,
a rental agreement and
a PhD regulations document.
In the development stage, we considered first 10 sentences of each document, based on which the rules and heuristics were defined.
For the evaluation, we used the next 10 sentences of each document.
The four documents used in our experiments are:
\begin{compactenum}
\item PhD regulations from Chalmers University
\item Equipment rental agreement from RSO, Inc.\footnote{\url{http://www.rsoinc.com/pdfs/equip_rental_revb.pdf}}
\item Terms of service for GitHub, Inc.\footnote{\url{https://help.github.com/articles/github-terms-of-service/}}
\item Terms of service for Facebook\footnote{\url{https://www.facebook.com/legal/terms}}
\end{compactenum}
After preparing the test sets and applying our tool, the tabular output from each set was evaluated manually according to the following criteria.

\subsection{Evaluation Criteria}

In our initial evaluation, we use a simple precision-recall metric over the following fields: Subject, Verb, Object and Modality.
The other fields (Time, Adverbials, Conditions and Notes) of our table structure are not included in the evaluation criteria as they are intrinsically too unstructured and will always require some post-editing in order to be formalised.
%Thus we thought it unfair to include them in the scores with the other fields.

\textbf{Precision} is concerned with rating the accuracy of the output of the tool.
For each value in the respective fields, we assign a point when it matches with our assessment of what the correct value should be.
When a single sentence results in a refinement with multiple clauses, we score each of these individually.

\textbf{Recall} is a measure of how much of the intended information the tool was able to extract.
For each sentence in the original text, we check whether the correct fields have been extracted by the tool, scoring accordingly.
When a sentence should result in multiple clauses, we score for each of these individually.

The local scores for precision and recall are often identical, because a sentence in the original text would correspond to one row (clause) in the table.
This is not the case when unnecessary refinements are added by the tool or, conversely, when coordinations in the text are not correctly added as refinements.

The evaluation was performed twice: first when using only the rules (Section~\ref{sec:rules}), and then again when using the rules and heuristics together (Section~\ref{sec:heuristics}).
A summary of our experimental results can be found in Table~\ref{tab:results}, including the harmonic mean scores ($F_1$) between precision and recall.

\begin{table}
  \centering
  \caption{Evaluation results based on a small set of test sentences (10 per document).}
  \label{tab:results}
  \medskip
  \begin{tabular}{l|ccc|ccc}
      \textbf{Document}
    & \multicolumn{3}{l|}{\bfseries Rules only}
    & \multicolumn{3}{l}{\bfseries Rules \& heuristics} \\
    & \textit{Precision}
    & \textit{Recall}
    & $F_1$
    & \textit{Precision}
    & \textit{Recall}
    & $F_1$ \\
    \hline
    PhD      & 0.66 & 0.73 & 0.69 & 0.82 & 0.90 & 0.86 \\
    Rental   & 0.75 & 0.67 & 0.71 & 0.71 & 0.66 & 0.69 \\
    GitHub   & 0.46 & 0.53 & 0.49 & 0.48 & 0.55 & 0.51 \\
    Facebook & 0.43 & 0.54 & 0.48 & 0.43 & 0.57 & 0.49
  \end{tabular}
\end{table}

\subsection{Observations} % Error analysis

The first observation from the results is that the $F_1$ score varies quite a lot between documents; from 0.49 to 0.86.
This is mainly due to the variations in language style present in the documents.
Overall the application of heuristics together with the rules does improve the scores obtained.%, though this is not equal across the different documents.

On the one hand, many of the sentence patterns which we handle in the heuristics appear only in the development set and not in the test set.
On the other hand, there are few cases which occur relatively frequently among the test examples but are not covered by the development set.
For instance, the introductory part of a sentence, the syntactic main clause, is sometimes pointless for our formalism, and it should be ignored, taking instead the sub-clause as the semantic main clause, e.g.:
\begin{quote}\itshape
User understands that ...
\end{quote}

The small corpus size is of course an issue here, therefore we cannot make any strong statements about the representative coverage of the development and test sets.

Analysing the modal verb \emph{shall} seems to be particularly difficult to get right.
It may either be an indication of an obligation when concerning an action, or it may be used as a prescriptive construct as in \emph{shall be} which is more indicative of a declaration.

\subsection{Paraphrasing}

In some sense, the task of extracting the correct fields from each sentence can be seen as paraphrasing from the given sentence into one of the known patterns which can be handled by our rules.
We give here some examples of errors encountered in the experiments, which can only currently be fixed by making non-trivial paraphrasing.

% e.g., ``You must be 13 years or older to use this Service.'' (GitHub) vs. ``You will not use Facebook if you are under 13.''

% 1

\begin{quote}\itshape
GitHub reserves the right at any time to modify or discontinue, temporarily or permanently, your access to the API with or without notice.
\end{quote}
For this sentence, our tool picks up \emph{reserve} as the verb and \emph{right} as the object, but this should really be realised as a permission with \emph{modify} as the verb and \emph{access to API} as the object.
This could furthermore be refined as a permission of a choice of actions (modify, discontinue temporarily, discontinue permanently).
Additional phrases such as \emph{at any time} and \emph{with or without notice} are actually not informative here, as they reflect the default behaviour of the formalism (i.e. lack of constraints).

% 2

\begin{quote}\itshape
We require applications to respect your privacy, and your agreement with that application will control how the application can use, store, and transfer that content and information.
\end{quote}
Here we get an obligation with the subject \emph{we}, the verb \emph{require} and the object \emph{applications to respect your privacy}.
The correct encoding however would be to make \emph{applications} the subject, with \emph{respect} as the verb and the object being \emph{your privacy}.

% 3

\begin{quote}\itshape
When you publish content or information using the Public setting, it means that you are allowing everyone, including people off of Facebook, to access and use that information, and to associate it with you.
\end{quote}
In this case, the tool fails completely, returning \emph{it} as the subject, \emph{mean} as the verb, and \emph{allowing everyone} as the object, with the declarative modality (D).
The entire \emph{when} clause should be treated as a condition.
The phrase \emph{you are allowing everyone} should more correctly be paraphrased as \emph{everyone is allowed}, making this actually a permission with the subject \emph{everyone}, the verb \emph{access} and the object \emph{that information}.

% 4

\begin{quote}\itshape
To learn more about Platform, including how you can control what information other people may share with applications, read our Data Policy and Platform Page.
\end{quote}
Sentences like this one are generally unimportant for our goals and should be ignored altogether, however we currently have no way identifying unhelpful sentences and ignoring them.

% 5 - TODO: passive to active voice

%% ---------------------------------------------------------------------------

\section{Formal Analysis}
\label{sec:analysis}

The ultimate goal of formalising normative texts is to be able to perform automated analysis, by which we mean running queries of various kinds against the \codiag\ model.

\textbf{Syntactic queries} are those which can be checked at a syntactic level, such as checking if a text contains any permissions for a particular agent or identifying obligations without constraints or reparations.
These are based on \emph{predicates} defined over single clauses, which serve as the building block for defining queries over the entire model.
The predicate $\textit{isObl}(C)$ for example is true if the clause $C$ is an obligation.
Syntactic queries are expressed by combining such predicates.

Queries which deal with timing constraints, possibility and invariance are called \textbf{semantic queries}.
For example, checking whether a clause could be enacted within a certain amount of time may depend on a previous sequence of events.
Determining whether these events may or may not happen cannot be done from the syntax alone.
Such queries are computed by converting the \codiag\ model into a network of timed automata and then applying the \UPPAAL{} model checker.
Semantic queries are expressed using \UPPAAL{}'s requirement specification language which is a subset of timed computation tree logic (TCTL).

%% ---------------------------------------------------------------------------

\section{Related Work}

% Information extraction is a large topic where examples generally tend to consist of applying standard NLP techniques combined with customised rules to some specific domain and problem.

Our work can be seen as similar to that of \citet{WynerPeters2011},
who present a system for identifying and extracting rules from legal texts using the Stanford parser and other NLP tools
% within the General Architecture for Text Engineering (GATE) system \cite{Cunnigham+2002}.
within the GATE system \cite{Cunnigham+2002}.
%the language of the source material so that they can be expressed using conditional and deontic rules.
% and translating into an executable logic
Their approach is somewhat more general, producing as output an annotated version of the original text.
Ours is a more specific application of such techniques, in that we have a well-defined output format which guided the design of our extraction tool,
which includes in particular the ability to define clauses using refinement.

% a passage from the US Code of Federal Regulations,
% US Food and Drug Administration, Department of Health and Human Services
% regulation for blood banks on testing requirements for communicable disease agents in
% human blood, Title 21 part 610 section 40.6 This is a four page document of 1,777 words.

% The paper presents a linguistically-oriented, rule-based approach
% It outlines use cases, discusses the source materials, reviews the methodology, then provides initial results and future steps.

\citet{Mercatali2005} tackle the automatic translation of textual representations of laws to a formal model, in their case UML.
This underlying formalism is of course different, where they are mainly interested in the hierarchical structure of the documents rather than the norms themselves.
Their method does not use dependency or phrase-structure trees but shallow syntactic chunks. %a variety of NLP info (tags, chunks, etc).

\citet{cheng2009} also describe a system for extracting structured information for texts in a specific legal domain.
Their method combines surface-level methods like tagging and named entity recognition (NER) with semantic analysis rules which were hand-crafted for their domain and output data format.

An alternative approach to bridging the gap between natural and formal languages is to introduce a \emph{controlled natural language} (CNL)---a reduced subset of a natural language which is in fact formal.
This may be done using either a general-purpose CNL such as Attempto Controlled English (ACE) \cite{fuchs:reasoningweb2008} which comes with a parser to discourse representation structures, or a custom CNL.

In their work, \citet{cnl2014} have designed a custom CNL which can be parsed directly into a \codiag.
The issue here is that the original natural language text may be quite a distance away from its CNL representation, and the translation from NL to CNL is still a manual step.

The FrameNet-CNL framework \cite{Barzdins14,Gruzitis-Dannells:2015:LRE} proposes an approach to information extraction problem by combining CNL with FrameNet---a lexicographic database describing word meanings based on the principles of frame semantics.
Such a system would encompass a powerful abstract knowledge representation paradigm along with a real-world information extraction system, based on frame-semantic parsing.
We consider the incorporation of semantic frames into our method as a direction for future work.

% TODO: "Formal Models of Sentences in Dutch Law" by Emile de Maat and Radboud Winkels - Page 28 (32) of http://wyner.info/research/Papers/AHLTL2011Papers.pdf

%% ---------------------------------------------------------------------------

\section{Conclusions and Future Work}

As already mentioned in the introduction and Section \ref{sec:analysis}, the main motivation of our work is to perform complex formal analyses of normative texts through syntactic and (timed) semantic queries through model checking.
We are in this paper helping to bridge the gap between natural language texts and formal analysis tools, taking the burden from the user to have to deal directly with formal languages.
%An implementation of \codiags\ into timed automata has been reported somewhere else \cite{Diaz2014}.
The approach presented here is thus not positioned as an eventual replacement for a human;
rather it is to be seen as an aid for a semi-automatic transition from natural language to \codiags.

%The current work is still in its early stages, and
Though the results of the experiments reported here are indicative at best, because of the small test corpus, the application of our technique to the case studies reported here has significantly increased the efficiency in the task of ``encoding'' the documents into \codiags.
Moreover, they are promising enough to warrant further work in this direction.

While the evaluation performed here measures the accuracy of the tool in terms of precision and recall,
another relevant metric would be to measure the time spent building a model from scratch versus the time spent on post-editing the output from our tool.

% Syntactic vs. semantic gap (syntactic vs. semantic transformation/paraphrasing).
It is common that some paraphrasing is needed during the post-editing phase.
This may be on the syntactic level, e.g. by fixing adverbial attachment.
However it may also require more in depth understanding, and often involves using related or opposite concepts which cannot be determined without more elaborated processing on the semantic level.

% Declarative sentences and states vs. actions.
The \codiag{} formalism is essentially \emph{action-based}, where clauses prescribe what an agent should or should not \emph{do}.
However in the texts from our experiments we have found that it is very common to describe what should or should not \emph{be}, i.e. referring to states of affairs.
Handling these kinds of sentences will require more effective paraphrasing patterns for such cases.

\subsubsection{Acknowledgements}
This research has been supported by the Swedish Research Council under Grant No. 2012-5746 (Reliable Multilingual Digital Communication: Methods and Applications) and partially supported by the Latvian State Research Programme NexIT (Project No. 1).

\bibliography{refs}

\end{document}